\documentclass[10pt,twocolumn,letterpaper]{article}
\usepackage[pagenumbers]{cvpr} 

\usepackage{amsmath}
\usepackage{amssymb}
\usepackage{mathtools}
\usepackage{amsthm}
\usepackage{enumitem}
\usepackage{arydshln}
\usepackage{booktabs}
\usepackage{colortbl}
\usepackage{graphicx}
\usepackage{arydshln}
\usepackage{amsfonts}
\usepackage{bbm}
\definecolor{mygray}{gray}{0.9}
\usepackage{algorithm}
\usepackage{listings}
\usepackage{graphicx}
\usepackage{makecell}
\usepackage{multirow}
\usepackage{booktabs}
\usepackage{adjustbox}
\usepackage{lineno}
\usepackage{etoolbox}
\theoremstyle{plain}
\newtheorem{theorem}{Theorem}[section]
\newtheorem{proposition}[theorem]{Proposition}

\theoremstyle{definition}

\theoremstyle{remark}

\usepackage{booktabs}
\usepackage{graphicx}
\definecolor{mygray}{gray}{0.5}
\usepackage{color}
\usepackage{colortbl}
\usepackage{array}
\usepackage{pifont}

\definecolor{mygray}{gray}{0.9}
\definecolor{mygreen}{RGB}{93,174,86}
\definecolor{darkgreen}{rgb}{0.0, 0.5, 0.0}
\definecolor{darkred}{rgb}{1.0, 0.0, 0.0}

\definecolor{cvprblue}{rgb}{0.21,0.49,0.74}
\usepackage[pagebackref,breaklinks,colorlinks,allcolors=cvprblue]{hyperref}

\def\confName{CVPR}
\def\confYear{2025}

\title{Seeing Far and Clearly: Mitigating Hallucinations in MLLMs \\ with Attention Causal Decoding}

\author{\normalsize Feilong Tang$^{1,2\ast}$, Chengzhi Liu$^{3\ast}$, Zhongxing Xu$^{1\ast}$, Ming Hu$^{1}$, Zelin Peng$^{4}$, Zhiwei Yang$^5$, \\ \normalsize Jionglong Su$^{3}$, Minquan Lin$^6$, Yifan Peng$^7$, Xuelian Cheng$^{1}$, Imran Razzak$^{2\dagger}$, Zongyuan Ge$^{1\dagger}$ \\ \\ \normalsize
$^1$Monash University,
$^2$MBZUAI,
$^3$XJTLU,
$^4$Shanghai Jiaotong University, \\ \normalsize
$^5$Fudan University,
$^6$University of Minnesota,
$^7$Cornell University,
 \\ \normalsize
{\small \texttt{Feilong.Tang@monash.edu}}}

\begin{document}
\maketitle
\renewcommand{\thefootnote}{\fnsymbol{footnote}}
\def\thefootnote{$^{*}$}\footnotetext{Equal contribution. {$\dag$} Corresponding authors. \\ Project Page: \href{https://mllms-farsight.github.io/}{\textit{https://mllms-farsight.github.io/}}}

\begin{abstract}
\quad Recent advancements in multimodal large language models (MLLMs) have significantly improved performance in visual question answering. However, they often suffer from hallucinations. In this work, hallucinations are categorized into two main types: initial hallucinations and snowball hallucinations. We argue that adequate contextual information can be extracted directly from the token interaction process. Inspired by causal inference in the decoding strategy, we propose to leverage causal masks to establish information propagation between multimodal tokens. The hypothesis is that insufficient interaction between those tokens may lead the model to rely on outlier tokens, overlooking dense and rich contextual cues. Therefore, we propose to intervene in the propagation process by tackling outlier tokens to enhance in-context inference. With this goal, we present FarSight, a versatile plug-and-play decoding strategy to reduce attention interference from outlier tokens merely by optimizing the causal mask. The heart of our method is effective token propagation. We design an attention register structure within the upper triangular matrix of the causal mask, dynamically allocating attention to capture attention diverted to outlier tokens. Moreover, a positional awareness encoding method with a diminishing masking rate is proposed, allowing the model to attend to further preceding tokens, especially for video sequence tasks. With extensive experiments, FarSight demonstrates significant hallucination-mitigating performance across different MLLMs on both image and video benchmarks, proving its effectiveness.
\end{abstract}
\vspace{-0.5cm}
\section{Introduction}
\label{sec:intro}

\begin{figure}[t]
\centering
\includegraphics[width=0.47\textwidth]{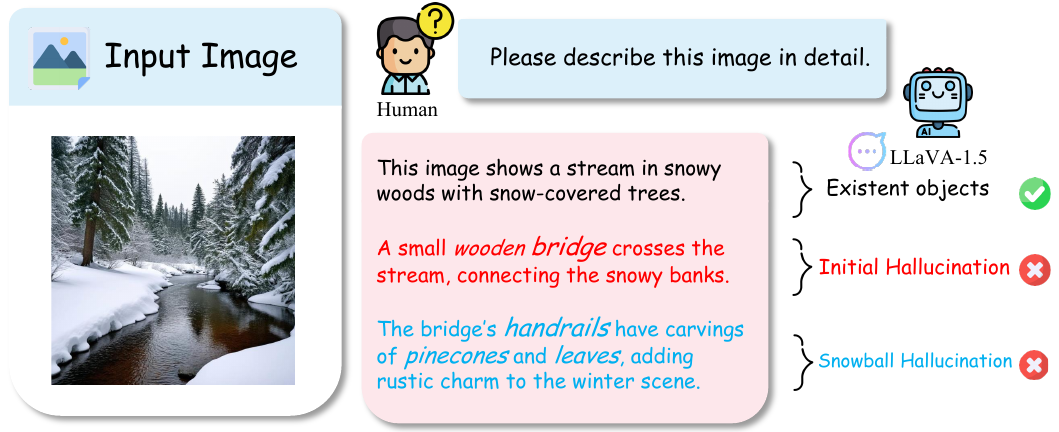}
\vspace{-0.4cm}
\caption{Illustrates the phenomenon of snowball hallucinations as an extension of initial hallucinations. MLLMs produce hallucinations by asserting nonexistent objects (\textit{e.g.,} \texttt{bridge}) within the image, followed by further explanatory errors (\textit{e.g.,} \texttt{handrails}). This progression from initial to snowball hallucinations reveals the model's tendency to build upon its own erroneous assumptions.}
\vspace{-0.2cm} 
\label{intro1}
\label{intro}
\end{figure}

Multimodal large language models (MLLMs)~\cite{zhu2023minigpt, zhang2023internlm, chen2023shikra, bai2023qwen, Instructblip, liu2024visual, liu2024improved, dong2024internlm} have become essential tools in addressing numerous vision tasks and performing complex visual question-answering due to their superior capabilities in content comprehension~\cite{lai2024lisa} and generation~\cite{geng2024instructdiffusion}. Despite their remarkable versatility, MLLMs often suffer from \textit{hallucinations}. Specifically, MLLMs frequently generate convincing text responses that contradict the visual content of an image, describing elements not present in the image. Hallucinations can be categorized into two types: initial hallucinations and snowball hallucinations, as illustrated in Fig.~\ref{intro}. Specifically, initial hallucinations (\textit{e.g.,} \texttt{bridge}) stem from insufficient information within the model, while snowball hallucinations (\textit{e.g.,} \texttt{handrails}) occur when the model maintains consistency with previous hallucinations.

\begin{figure}[t]
\centering
\begin{tabular}{cc}
\includegraphics[width=0.47\textwidth]{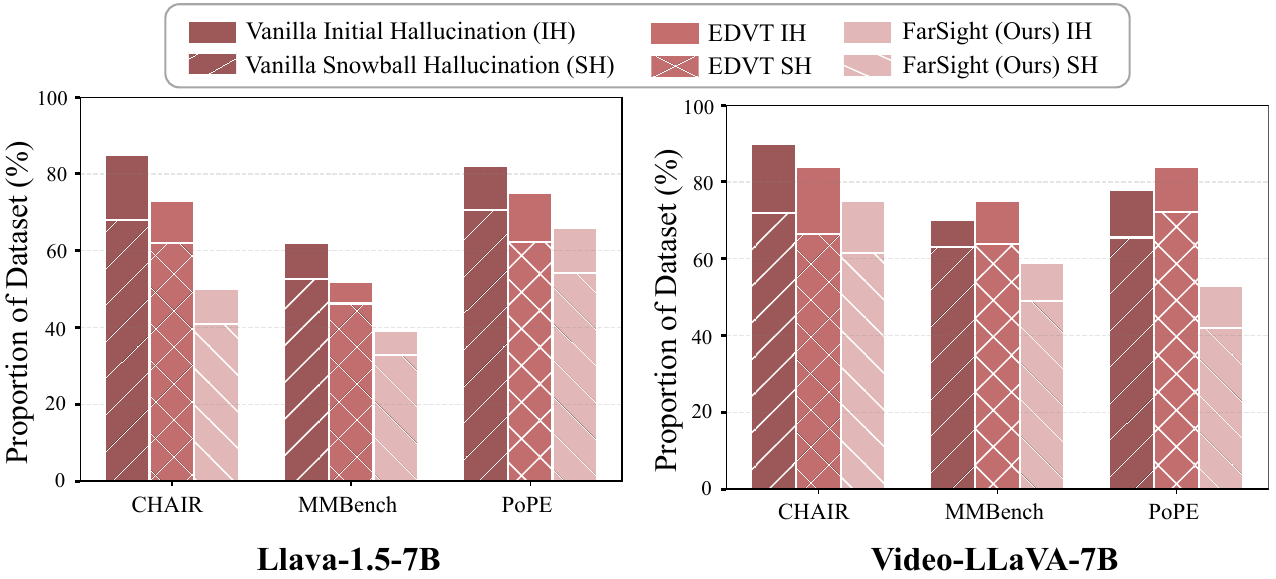}
\end{tabular}
\vspace{-0.3cm}
\caption{Percentage of initial hallucination (IH) and percentage of snowball hallucination (SH) (calculated over the entire datasets) for LLaVA-1.5-7B~\cite{liu2024improved}, Video-LLaVA-7B~\cite{lin2023video} and EDVT~\cite{Ma_2024_CVPR}.}
\vspace{-0.5cm}
\label{intro2}
\end{figure}

The key to mitigating hallucinations lies in extracting contextual information from the token interaction process. Recent studies focus on external knowledge retrieval~\cite{shuster2021retrieval,caffagni2024wiki} and robust instruction fine-tuning~\cite{yu2024rlhf,sarkar2024mitigating,xiao2024detecting}, but these methods often incur substantial additional costs. Conversely, other approaches focus on training-free decoding strategies such as contrastive decoding~\cite{leng2024mitigating,wang2024mitigating,huo2024self,kim2024instructive} and self-calibrating attention~\cite{huang2024opera,liu2024paying,xing2024mitigating,ma2024vista}. They aimed to enhance the accuracy and consistency of generated responses by reducing excessive reliance on linguistic priors in the token interaction process. Though previous works have shown effectiveness, they lack analysis of the interaction process between multimodal tokens and the causes of hallucinations. For example, Fig.~\ref{intro2} illustrates a high proportion of snowball hallucinations, particularly in video captioning. Interestingly, these methods have not been effective in reducing the proportion of snowball hallucinations. In this study, we hypothesize that insufficient interaction between tokens may result in over-reliance on outlier tokens, thereby neglecting dense and informative contextual cues. In this work, we argue that intervening effectively in the token interaction process enhances in-context inference. 
Moreover, existing causal mask refinements (\textit{e.g.,} ALiBi~\cite{press2021train}, StableMask~\cite{yin2024stablemask}, T5~\cite{raffel2020exploring}) primarily improve token interactions and target unimodal text extrapolation. In contrast, our FarSight explicitly addresses multimodal hallucinations by enhancing vision-language token interactions in MLLMs.

\begin{figure}[t]
\centering
\begin{tabular}{cc}
\includegraphics[width=0.45\textwidth]{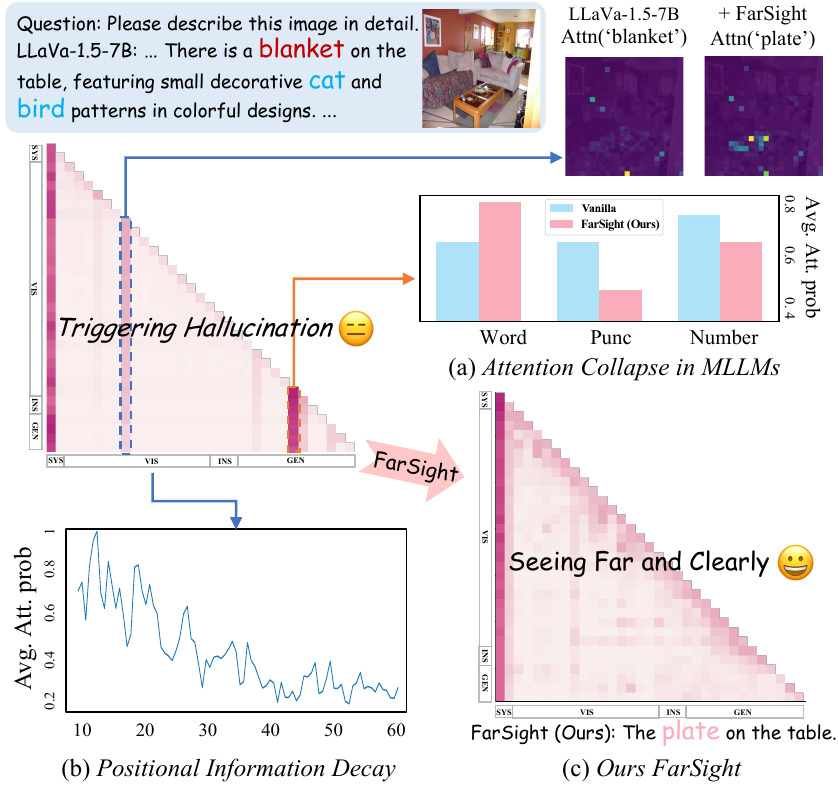}
\end{tabular}
\vspace{-0.3cm}
\caption{(a): Attention Collapse in MLLMs: Outlier tokens from different modalities are assigned disproportionately high attention scores, hindering interaction between relevant tokens. (b): Positional Information Decay: As text generation progresses, attention to visual information gradually diminishes. (c): Our FarSight, as a plug-in, mitigates these issues by effectively reducing attention interference from outlier tokens and improving response accuracy.}
\vspace{-0.5cm}
\label{intro3}
\end{figure}

To delve deeper into this phenomenon, we analyze the attention maps during decoding and identify two issues contributing to hallucinations.  \textit{(i) Attention Collapse in MLLMs}: As illustrated in Fig.~\ref{intro3} (a), we observe that the model tends to allocate disproportionate attention to tokens with limited informational content. These low-information yet high-attention outlier tokens, such as visual backgrounds and textual symbols, disrupt the effective propagation of relevant information. This issue arises because the softmax attention mechanism requires all attention scores to be non-zero and sum to one, causing even low-information or non-priority tokens to receive disproportionate attention. Attention collapse, akin to the findings in Opera~\cite{huang2024opera} on the ``summary token", causing a gradual attenuation of vision and text information transmission as the generated text extends. \textit{(ii) Positional Information Decay}: As illustrated in Fig.~\ref{intro3} (b), we observe a progressive decline in attention to dense vision information throughout the generation process. This occurs due to the rotational position encoding (RoPE)~\cite{su2021roformer}, whose long-term decay fails to provide adequate positional information to ensure sufficient interaction between vision and text tokens. As the relative distance increases, the flow of vision token information gradually diminishes, leading to potential hallucinations. Therefore, \textit{our findings indicate that maintaining balanced information propagation and refining positional encoding can mitigate attention collapse and positional information decay, both of which contribute to hallucinations.}

In this work, we propose FarSight, a versatile plug-and-play decoding strategy that reduces attention interference from outlier tokens by optimizing the causal mask. Specifically, we initialize a set of attention registers within the upper triangular matrix of the causal mask to capture attention diverted to outlier tokens. These attention registers retain the causal decoding properties, ensuring that information from future tokens is not accessed prematurely. Additionally, we design a dynamic register-attention distribution mechanism that explicitly optimizes attention allocation at each decoding step for robust in-context inference. 

The core of our method is to optimize the effective propagation of tokens. We modulate attention distribution for tokens with multimodal informational content to improve token propagation. Furthermore, the relative positional limitations of RoPE encoding lead to insufficient transmission of vision-to-text token information during contextual interactions, which undermines positional awareness. Therefore, we introduce a progressively diminishing masking rate within the causal mask to encode absolute positional information, allowing the model to attend to further distant preceding tokens, especially for video sequence tasks.

With extensive experiments, FarSight demonstrates significant hallucination-mitigating performance across different MLLMs on both image and video benchmarks, proving its effectiveness. Our contributions are as follows:
\begin{itemize}
    \item We analyze the self-attention token propagation patterns, revealing two main causes of hallucinations in MLLMs: attention collapse and positional information decay.
    \item We propose FarSight, a plug-and-play decoding strategy that effectively mitigates hallucinations stemming from these issues by merely adjusting the causal mask.    
    \item Extensive evaluations on both image and video tasks demonstrate the superior performance of FarSight, offering an effective solution for mitigating hallucinations.
\end{itemize}

\section{Related Work}
\paragraph{Hallucinations in MLLMs.} Leveraging open-source large language models like LLaMA~\cite{touvron2023llama,touvron2023llama2} and Vicuna~\cite{chiang2023vicuna}, MLLMs~\cite{huang2023diversity,chen2023sharegpt4v,young2024yi,elhoushi2024layer,ye2024mplug,li2024mini,internlmxcomposer2_5,li2024llava,tong2024cambrian,mckinzie2024mm1,zhang2024mm1} can understand and generate a wide range of content more effectively by combining information from multiple modalities, such as text, images, and audio. Hallucination in MLLMs~\cite{ji2023survey,tang2025intervening,liu2024survey,hu2023ciem,liu2023aligning,zhang2023siren,ghosh2024exploring,ghosh2024medsumm,sahoo2024enhancing,liu2024phd} refers to the generation of text that is misaligned with the content of the provided images. Hallucination may originate from reliance on model priors~\cite{factuality,gunjal2024detecting,liu2023mitigating,zhao2023beyond,zhai2023halle,li2024monkey,jain2024vcoder,cho2022fine}, limited knowledge comprehension~\cite{manakul2023selfcheckgpt,zhou2023analyzing,hu2023ciem,liu2023aligning,you2023ferret}, or an inability to effectively contextualize the given input~\cite{li2022contrastive,chuang2023dola,liu2024paying,huang2023survey,wang2023evaluation,liu2024seeing,xue2025mmrc}. According to the causes of hallucination, hallucinations can be classified into two types: initial hallucinations~\cite{ouyang2022training,wang2022self} occur due to the model lacking necessary information; snowball hallucinations~\cite{zhong2024investigating,zhang2023language} arise when the model generates a series of hallucinations to maintain consistency with previous ones, even when the required knowledge is available. In this paper, we primarily conducted experiments and analyses on image and video benchmarks.

\noindent \textbf{Hallucination Mitigation for MLLMs.} Researchers have proposed various strategies, from data optimization to model adjustments, to improve the accuracy and consistency of generated content. To mitigate hallucination, solutions include robust instruction tuning~\cite{jiang2024hallucination,yu2024rlhf,yu2024hallucidoctor,yue2024less}, post-hoc processing using auxiliary analysis networks~\cite{yin2023woodpecker,logical,collaboration,zhang2024recognize}, and various decoding strategies~\cite{leng2024mitigating,favero2024multi,wang2024mitigating,kim2024instructive,ibd,huang2024opera}. Recent studies have focused on outlier tokens, causing generated text to emphasize summarizing information from these tokens rather than utilizing dense and rich contextual cues. Additionally, some studies~\cite{ma2024vista,xing2024mitigating} have found that RoPE positional encoding is insufficient to support information propagation between multimodal tokens in contextual reasoning. Moreover, existing causal mask refinements (\textit{e.g.,} ALiBi~\cite{press2021train}, StableMask~\cite{yin2024stablemask}, T5~\cite{raffel2020exploring}) primarily improve token interactions and target unimodal text extrapolation. This paper proposes an optimized causal masking approach to extract sufficient contextual information during token interactions, effectively mitigating hallucination without additional training, data, or inference time.

\section{Preliminary and Motivation}
\subsection{Paradigm of MLLMs Generation}\label{sec3.1}

\textbf{Vision and Language Inputs.} The inputs of MLLMs consist of both image and text. Generally, the raw images are commonly fed to the visual encoder. Then the cross-model projection module maps vision information into LLMs’ input space, which is denoted as vision tokens $\mathbf{x}^v = \{x_0,x_1,\ldots,x_{N-1}\}$ where $N$ is the length of vision tokens. Similarly, text is processed by tokenizer and embedding modules, which is denoted as text tokens $\mathbf{x}^t = \{x_N,x_{N+1},\ldots,x_{M+N-1}\}$ where $M$ is length of text tokens. Then, the image and text tokens are concatenated as the final input and denoted as $\{x\}_{t=0}^{T-1}$ where $T=N+M$.

\noindent\textbf{MLLMs Forward.} The backbone networks of MLLMs $M_{\theta}$ are pre-trained LLMs (\textit{e.g.,} Vicuna~\cite{chiang2023vicuna} and LLaMA 2~\cite{chiang2023vicuna}), parameterized by $\theta$ that auto-regressively generates responses. Given a multimodal input sequence $\mathbf{x}$, the model maps the logit distribution to the next token prediction output $y_t\in\mathbb{R^{\left\vert\mathcal{V}\right\vert}}$ at time step $t$ in the vocabulary set $\mathcal{V}$:
\begin{equation}
y_t \sim p_{\theta}(y_{t}|\mathbf{x},y_{<t}) \propto \text{logit}_{\theta}(y_{t}|\mathbf{x},y_{<t}),
\label{eq1}
\end{equation}
where $y_{<t}$ denotes all previously generated tokens $\{x_i\}_{i=0}^{t-1}$.
\vspace{-0.6cm}
\paragraph{Next Token Decoding.} 

After obtaining the next token probability $p(y_{t}|\mathbf{x},y_{<t})$, different decoding strategies~\cite{graves2012sequence, chuang2023dola, huang2024opera} are proposed to predict the next token. The decoded token is concatenated to the last of the original input text for the next-round generation, until the generation is ended.

\subsection{What Causes Hallucinations}\label{32se}
\paragraph{Attention Collapse in MLLMs.} 

We investigate the self-attention in the transformer block~\citep{transformer} of the auto-regressive decoder and leverage a column-wise product to calculate metric values. Denote the current generated sequence as $\{x_i\}_{i=0}^{t-1}$ and their causal self-attention weights as $\{\omega_{t-1,j}\}_{j=0}^{t-1}$ applied to the next token prediction. The weights  $\omega \in \mathbb{R}^{n \times n}$ can be obtained from the softmax function as follows:
\begin{equation}
\mathcal{O} = \text{SoftMax}(\omega)\cdot V, \quad \omega=\frac{Q \cdot K^\top}{\sqrt{d_{l}}} + M,
\label{eq:causal_softmax}
\end{equation}
where $Q, K, V \in \mathbb{R}^{n \times d_l}$ are the Query, Key, and Value matrices. $n$ and $d_l$ are the sequence length and the hidden dimensions, $M \in \mathbb{R}^{n \times n}$ is the causal mask, and $\mathcal{O}$ is the output. The causal mask $M$ ensures that the model does not attend to future tokens, preserving causality in the sequence. The attention weights are structured as follows:
\begin{equation}
    \omega_{i} = [\omega_{i1}, \omega_{i2}, \cdots, \omega_{ii}, 0, \cdots, 0]_n.
\end{equation}

\begin{proposition}[Attention Collapse in MLLMs] Let inputs be sampled from a data distribution $q(x_1, x_2, \dots, x_N)$ and processed by a contextual, layer-wise decoder with attention layers. Define the disproportionality in an attention layer as measured by the total probability of prefixes $\sum_{x_{<N}} q(x_{<N})$, where the attention collapse after applying softmax for $i < n < N$ in the $l$-th layer satisfies:
\begin{equation}
\sum^{N}_{n=1} \sum_{j\leq i} \omega_{n,j}^{l} > \frac{I(x_{\leq i};x_{n+1})}{I(x_{\leq n};x_{n+1})}\sum^{N}_{n=1}\sum_{j\leq n} \omega_{n,j}^{l} + o(1),
\end{equation}
Here, $I(A;B)$ denotes the mutual information between two variables $A$ and $B$, indicating the amount of shared information between them. $I(x_{\leq n};x_{n+1})>0$ represents that the token $x_{n+1}$ is informationally dependent on the preceding sequence $x_{\leq n}$, quantifying how much information about $x_{n+1}$ is contained within $x_{\leq n}$.
\label{AC}
\end{proposition}
\noindent \textbf{Remark:} Proposition~\ref{AC} indicates that Attention Collapse refers to the phenomenon where the attention weights for certain tokens far exceed the informational contribution of those tokens. This often occurs with semantically irrelevant tokens, such as non-functional words (\textit{i,e.,} punctuation marks) and background vision tokens. As a result, the focus of the model diffuses across these irrelevant tokens, increasing perplexity during length extrapolation and hindering interaction among semantic tokens, as illustrated in Fig.~\ref{intro3} (a).
\\ \hspace*{\fill} \\
\noindent\textbf{Positional Information Decay.} The vanilla attention model lacks positional awareness, as it does not encode relative distance between tokens. In contrast, RoPE~\cite{su2021roformer} addresses this by encoding the positional data of tokens using a rotation matrix, which inherently includes an explicit relative position dependency. Within each attention $\omega$, RoPE is applied across all projected query $Q$ and key $K$ inputs to compute the attention weights by leveraging relative distance between tokens. Consequently, the attention with relative position embedding is expressed as:
\begin{equation}
\tilde{\omega}_{ij} = \frac{{R}_i\cdot {q}_i \cdot {R}^{T}_j \cdot {k}^{T}_j}{\sqrt{d_{l}}} = \frac{{q}_i \cdot {R}_{j-i} \cdot {k}^T_j}{\sqrt{d}_l},
\label{55}
\end{equation} 
where $R \in \mathbb{R}^{n \times n}$ denotes the rotary position embedding matrices applied to the query and key. $j-i$ stands for relative position between ${q}_i$ and ${k}_j$. The long-term decay refers to the decrease of $\tilde{\omega}_{ij}$ as the relative distance $j-i$ increases.

\noindent \textbf{Remark:} RoPE integrates relative position data by multiplying rotation matrices rather than appending positional embeddings to the input. The relative proximity between two tokens effectively determines their influence, as closer tokens should impact each other more than distant ones. However, using the same attention mechanism for both vision and text tokens results in unintentional text generation in MLLMs, as illustrated in Fig.~\ref{intro3} (b). Consequently, we argue that RoPE long-term decay limits multimodal tokens' information propagation, which contributes to hallucination. In contrast, maintaining absolute positional focus in generated text could allow the model to achieve precise positional awareness and improve response accuracy.

\section{Methodology}
\label{sec:FarSight}

\begin{figure}
    \centering
    \includegraphics[width=0.48\textwidth]{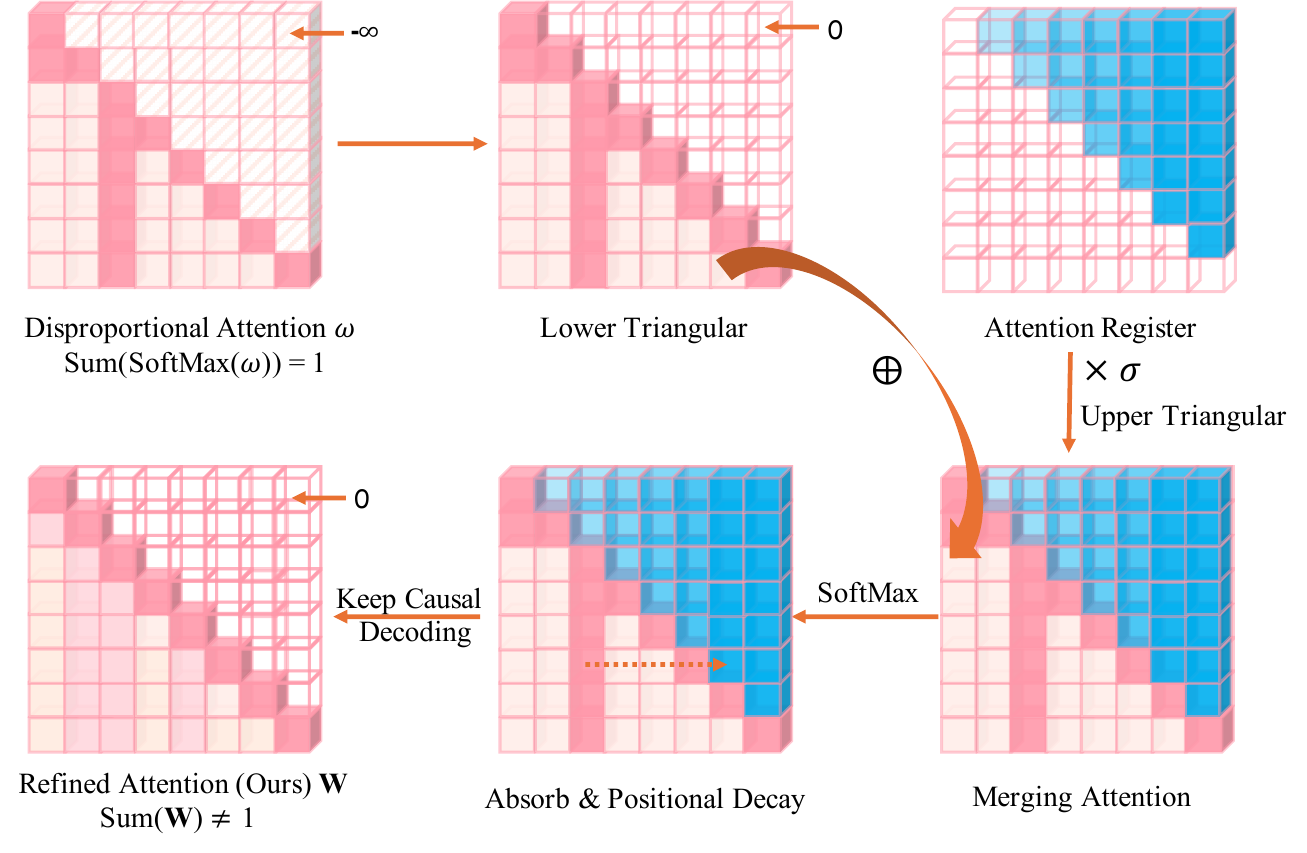}
    \vspace{-0.6cm}
    \caption{The scheme of the proposed FarSight strategy, which integrates with the softmax operation, replacing the traditional causal mask. Specifically, the attention score matrix $\omega$ is cleared of attention values in the upper triangular part, then register-attention scores are added using the matrix \(\mathcal{P}\) followed by the softmax computation. $\mathcal{P}$ has linear decay in the upper triangular part and zeros in the lower triangular part. After the softmax operation, the remaining attention probabilities in the upper triangular part are cleared  to ensure the causal decoding property is preserved.}
    \vspace{-0.3cm}
    \label{fig:solve}
\end{figure}

Fig.~\ref{fig:solve} provides an overview of the proposed strategy, built upon an LLM decoding paradigm in Section~\ref{sec3.1}. Attention registers, detailed in Section~\ref{sec41}, are introduced to absorb outlier tokens' attention scores, dynamically guiding the model toward contextually rich semantic information. Meanwhile, a progressively diminishing masking rate is introduced to capture absolute positional focus with rigorous theoretical justification, which is described in Section~\ref{sec42}. For ease of comprehension of how FarSight works, \cref{alg:code} exhibits the pseudo-code in the decoder layer. FarSight builds upon recent causal masking strategies~\cite{yin2024stablemask,press2021train} with a fully dynamic attention register mechanism tailored for vision-language token interactions in MLLMs.

\subsection{Upper Triangular Metric as Attention Registers}\label{sec41}

To alleviate attention collapse issues, we propose the dedicated attention register to allocate excess attention scores. For each $\omega$, we construct an upper triangular score matrix $\mathcal{P} \in \mathbb{R}^{n \times n}$ as attention register, defined as follows:
\begin{eqnarray}
\mathcal{P}_{i} = [\underbrace{0, 0, \cdots, 0,}_{i} \underbrace{\mathcal{P}_{i, i+1}, \mathcal{P}_{i, i+2}, \cdots, \mathcal{P}_{i,n}}_{n - i}]_n,
\vspace{-0.2cm}
\end{eqnarray}
where $\mathcal{P}_{i}$ allocates $n-i$ register-attention scores in each row to handle excess attention values while maintaining zero values for positions up to $i$.
To integrate $\mathcal{P}$ with $\omega$, we adjust $\omega$ by adding the register-attention scores from $\mathcal{P}$ as follows: 
\begin{equation}
    \mathbf{W} = \omega \cdot C + \mathcal{P}, \quad \text{where} \quad C = \text{tril}(\mathbbm{1}_{n \times n}),
\end{equation}
where $C \in \mathbb{R}^{n \times n}$ denotes a lower-triangular matrix filled with ones to ensure causal masking by allowing attention only to preceding or current tokens, as illustrated in Fig.~\ref{fig:solve}.

Since the model is training-free, the attention-registers $\mathcal{P}$ should not interfere with the original attention score distribution $\omega$ during inference and align with the relative positional encoding $R$ in Eq.~\ref{55} to maintain coherence in generated text. For $\mathcal{P}_{i,j}$, the values are defined as:
\begin{equation}
\mathcal{P}_{i,j} = - (j - i) \cdot \sigma, \quad \forall j > i,
\label{eq:decayfactor}
\end{equation}
where $\sigma$ is a decay rate hyperparameter. This setup ensures that $\mathcal{P}$ conforms to the gradual attenuation pattern in attention. 
Thus, the final attention score matrix with FarSight is defined as:
\begin{equation*}
\mathbf{W}_{i} = [\mathcal{P}_{i,1}, \mathcal{P}_{i,2}, \cdots, \mathcal{P}_{i,i}, -\sigma, -2\sigma, \cdots, -(n - i)\sigma]_n,
\end{equation*}
where \(\mathcal{P}_{i,j}\) denotes the original attention score at \((i,j)\), with \(\mathcal{P}_{i,i}\) capturing the self-attention along the diagonal. The decay factor \(\sigma \cdot (j - i)\) applied for future tokens \(i < j\), which enforces causal masking. The standard causal mask operation in Eq.~\ref{eq:causal_softmax} is then modified as:
\begin{equation}
\tilde{\mathbf{W}} = \text{SoftMax}(\underbrace{\omega \cdot C + \mathcal{P}}_{\mathbf{W}}) \cdot C.
\label{eq:remask}
\end{equation}
\noindent \textbf{Remark:} The \(\mathbf{W} = \omega \cdot C + \mathcal{P}\) within the \(\text{SoftMax}\) function incorporates register-attention scores by masking the attention matrix, while the \(C\) outside \(\text{SoftMax}\) ensures that any masked scores are reset to zero. This design enables FarSight to retain causal decoding properties, preventing information from future tokens is not accessed prematurely. The register-attention matrix \(\mathcal{P}\) effectively captures and buffers excess attention by providing dedicated slots for surplus values, ensuring that the main attention mechanism remains focused on relevant tokens without being distracted by irrelevant or future positions.

\subsection{Positional Awareness Encoding}\label{sec42}
The core idea of absolute position encoding is to modify the attention matrix so that the sum of actual attention scores (located in the lower triangular part of the attention matrix $\omega$) is not constrained to equal 1, as illustrated in Fig.~\ref{fig:solve}. Specifically, we introduce a progressively diminishing masking rate in the causal mask, allowing attention distributions to vary across positions, thereby effectively incorporating absolute positional information.

Let $\omega_i$ denote the raw attention scores in the $i$-th row, and let $\mathcal{P}_i$ denote the corresponding register-attention scores. Instead of a single softmax normalization over the entire row, we partition the normalization into two segments. For positions $j \le i$, the normalized contribution is defined as:
\[
\alpha_i(j)=\text{SoftMax}(\mathbf{W}),\quad j\le i,
\]
and for tokens at positions $j > i$, the normalized register-attention contribution is given by
\[
\gamma_i(j)=\text{SoftMax}(\mathbf{W}),\quad j> i.
\]
The model encodes positional information for a sequence of identical input tokens, $\mathbf{x} = \{x_i\}_{i=1}^{n} \in \mathbb{R}^n$, by leveraging both attention score accumulation and decay. Specifically, the actual attention scores \(\omega_{i,j}\) are uniform across each row. Consequently, the cumulative sum of their exponentiated values progressively increases with the row index $i$. This cumulative increase emphasizes information before the current position, contributing to the encoding of absolute positional information. Simultaneously, the cumulative sum of the exponentiated register-attention scores decreases as $i$ increases due to the applied decay in $\mathcal{P}_{i,j}$. This decay constrains attention on content after the current position, ensuring that attention primarily emphasizes preceding information. Consequently, we obtain:

\begin{algorithm}[t]
\caption{Pseudocode of FarSight in PyTorch Style.}
\label{alg:code}
\definecolor{codeblue}{rgb}{0.25,0.5,0.5}
\lstset{
  backgroundcolor=\color{white},
  basicstyle=\fontsize{6.5pt}{6.5pt}\ttfamily\selectfont,
  columns=fullflexible,
  breaklines=true,
  captionpos=b,
  linewidth=0.48\textwidth,
  commentstyle=\fontsize{7.2pt}{7.2pt}\color{codeblue},
  keywordstyle=\fontsize{7.2pt}{7.2pt},
  morekeywords={register_score},
}
\begin{lstlisting}[language=python]
# x: hidden input in each attention layer
# C: upper-triangular matrix filled with 1
# Sigma: decay factor, n_head: attention head

def register_score(self, seq_len: int):
    # Create a register (upper-triangular matrix with 0)
    register = 1 - torch.triu(torch.full((seq_len, seq_len), 1), diagonal=1)
    
    # Generate register alibi biases 
    register_score = get_alibi_biases(n_heads, -register.flip(dims=[1])).flip(dims=[1])
    
    # Final register score adjustment
    return register_score.contiguous() * (1 - mask)

def FarSightAttention(self, x: torch.Tensor):
    # query, key, value projection
    xq, xk, xv = qkv_proj(x) 
    
    #query, key, value projection and get QK^T/sqrt(d)
    scores = torch.matmul(xq, xk.transpose(2, 3)) / math.sqrt(self.hid_dim)
    
    # add register scores and introduce decay factor 
    scores = scores * C * sigma + register_score
    
    # remove register score to keep causal decoding
    scores = torch.softmax(scores, dim=-1) * C

    # final projection and output
    return self.wo(torch.matmul(scores, xv)) 
\end{lstlisting}
\end{algorithm}

\[
\sum_{j=1}^{i}\alpha_i(j) < \sum_{j=1}^{i+1}\alpha_{i+1}(j),
\]
indicating that, after applying Eq.~\ref{eq:remask}, the accumulated attention over valid tokens exhibits a monotonically increasing trend with respect to the row index \(i\), \ie $\tilde{\mathbf{W}} \cdot V = \sum_{i=1}^{n} \beta_i \boldsymbol{v}_i$, satisfying $\beta_1 < \beta_2 < \dots < \beta_n = 1$, progressively encoding the absolute positional context. This progressive allocation allows the model to maintain an ordered information flow across positions, where tokens at later positions aggregate increasingly more historical context from preceding tokens. As \( i \) grows, the model sharpens its focus on earlier tokens, reinforcing long-range dependencies and enhancing positional awareness in the generated sequence.

\begin{table}[t!]
\centering
\caption{Comparison of our \textbf{Positional Awareness Encoding} with other methods on the CHAIR~\cite{rohrbach2018object} and POPE~\cite{li-etal-2023-evaluating} datasets. RoPE: rotary positional embedding for both visual and text tokens, as used in the original MLLMs. FixVPE: fixed rotary embedding for visual tokens only. EDVT: rotary embedding for text tokens only.
}
\vspace{-0.2cm}
\resizebox{0.48\textwidth}{!}{
\begin{tabular}{l||cccc}
\midrule
\rowcolor[gray]{0.9} 
\textbf{Method} & \textbf{CHAIR$_S$ $\downarrow$} & \textbf{CHAIR$_I$} $\downarrow$ & \textbf{POPE-R} $\uparrow$& \textbf{POPE-P} $\uparrow$\\
\midrule
\midrule
LLaVA-1.5 (RoPE)& 48.0 & 13.9 & 87.0 & 82.8 \\
+ FixVPE & 47.3 & 13.4 & 87.5 & 84.7\\
+ EDVT & 46.8 & 14.5 & 87.8 & 85.4 \\
\rowcolor{gray!20} + \textbf{FarSight (Ours)} & 41.6 ~\color{blue!60}{({+6.4})}& 13.2 ~\color{blue!60}{({+0.7})}& 90.5~\color{blue!60}{({+3.5})} & 86.1~\color{blue!60}{({+3.3})} \\
\midrule
Video-LLaVA (RoPE) & 50.2 & 15.6 & 81.6 & 85.3 \\
+ FixVPE & 48.5 & 14.9 & 81.9 & 85.2 \\
+ EDVT & 46.8 & 13.7 & 82.5 & 84.7 \\
\rowcolor{gray!20} + \textbf{FarSight (Ours)} & 44.8~\color{blue!60}{({+5.4})}& 12.9~\color{blue!60}{({+2.7})} & 83.2~\color{blue!60}{({+1.6})} & 85.8~\color{blue!60}{({+0.5})} \\
\midrule
\end{tabular}
}
\label{Position}
\end{table}

\section{Experiments}
\label{sec:exp}

\subsection{Experimental Setup}\label{51}
\noindent\textbf{Baseline.}  
We select six representative MLLMs to evaluate performance across image and video tasks, including InstructBLIP~\cite{dai2023instructblipgeneralpurposevisionlanguagemodels}, LLaVA-1.5~\cite{liu2024improved}, VILA~\cite{lin2023vila}, Video-LLaMA2~\cite{damonlpsg2024videollama2}, Chat-UniVi~\cite{jin2024chat}, and Video-LLaVA~\cite{lin2023video}. InstructBLIP and LLaVA-1.5 primarily focus on image tasks, while VILA and Video-LLaMA2 specialize in video tasks. Chat-UniVi and Video-LLaVA are capable of processing both image and video data, allowing for a comprehensive evaluation across both modalities. More detailed descriptions are provided in Appendix A.

\noindent\textbf{Evaluation Benchmarks.} 
We conduct evaluations on both image and video benchmarks. For image benchmarks, we assess three categories: (1) Comprehensive benchmarks (MMBench~\cite{liu2025mmbench}, LLaVA$^{\mathrm{W}}$~\cite{liu2024visual}, MM-Vet~\cite{yu2023mm}); (2) General VQA benchmarks (VizWiz~\cite{gurari2018vizwiz}, SQA~\cite{lu2022learn}); (3) Hallucination benchmarks (POPE~\cite{li-etal-2023-evaluating}, CHAIR~\cite{rohrbach2018object}). For video, we evaluate three zero-shot video understanding datasets: MSRVTT-QA~\cite{kim2023provable}, MSVD-QA~\cite{Xu2017VideoQA}, and ActivityNet-QA~\cite{Zhou2017TowardsAL}, along with the Video-Based Text Generation Benchmark for quantitative analysis~\cite{maaz2023video}.

\noindent\textbf{Implementation Details.}
FarSight supports Greedy, Sampling, and Beam Search decoding strategies, with Greedy decoding used for illustration. Details of the other methods are in Appendix E. For the Decay Factor, we set the sequence length $(seq)$ to 256 and define the decay rate $\sigma$ in Eq.~\ref{eq:decayfactor} as $log_{\alpha}(seq)$, with $\alpha$ is 1024, the typical maximum token limit. Extensive experiments confirm $seq = 256$ ensure stable and consistent generation.

\begin{figure}
    \centering
\includegraphics[width=\linewidth, height=3.24cm]{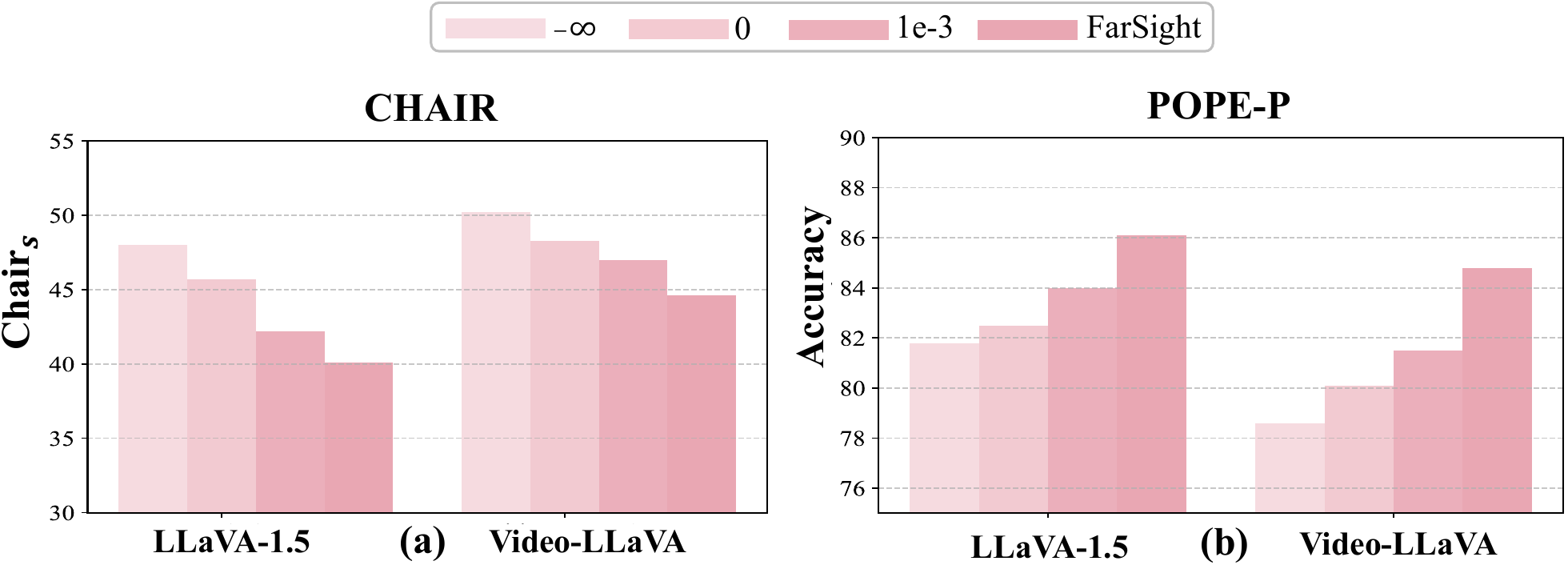}
    \caption{Comparison of different \textbf{Upper Triangular Attention Values in Attention Registers.} (a) and (b) show model performance with varying upper triangular attention values on the CHAIR and POPE-P datasets.}
\label{ablation}
\end{figure}

\begin{figure}
    \centering
    \includegraphics[width=\linewidth]{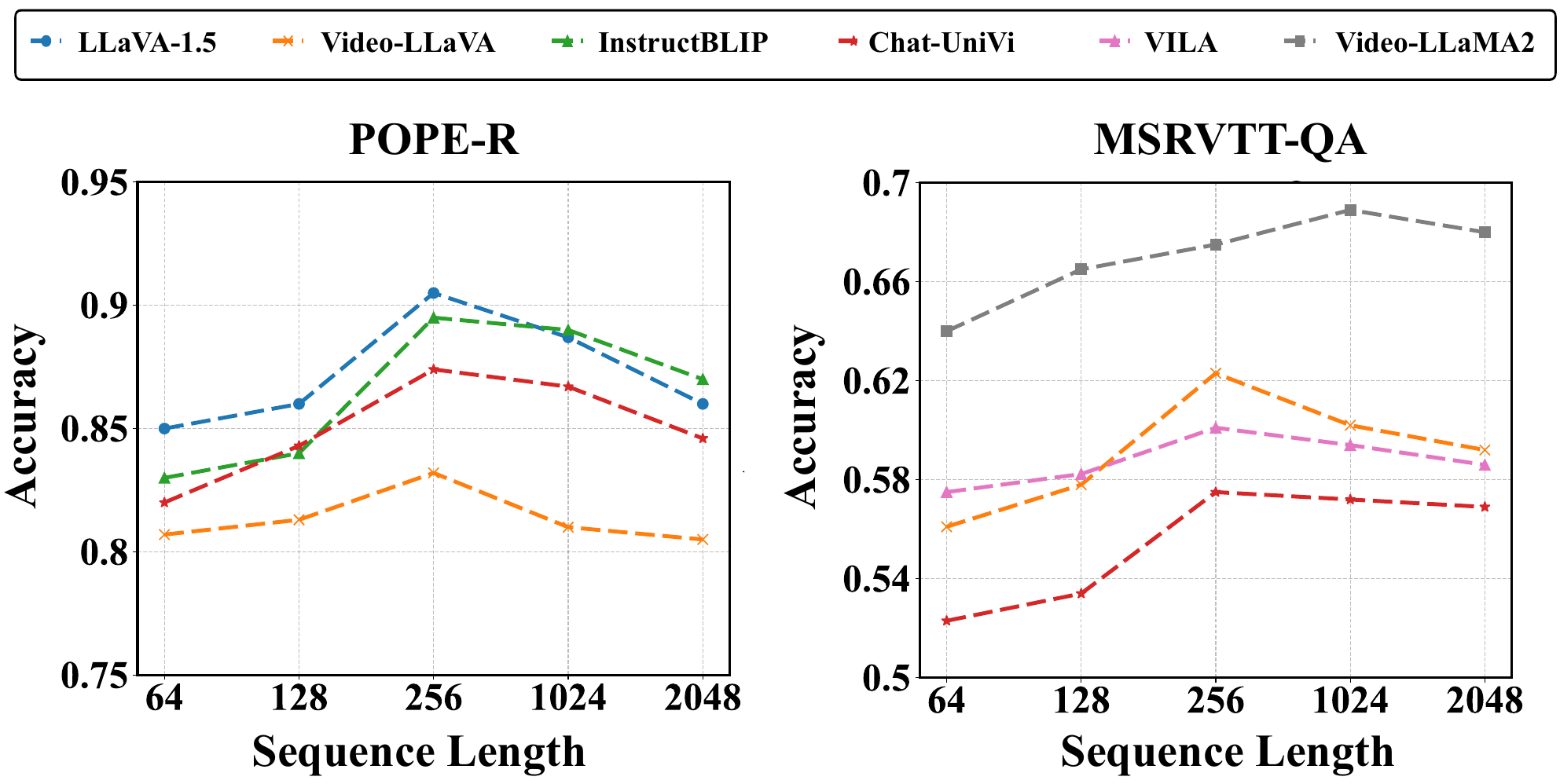}
    \caption{The impact of sequence length on attention decay and the performance of MLLMs on the POPE-R and MSRVTT-QA datasets integrated with our FarSight.}
\label{factor}
\vspace{-0.4cm}
\end{figure}

\begin{figure*}[!t]
  \centering
  \begin{tabular}{cc}
\includegraphics[width=0.98\linewidth]{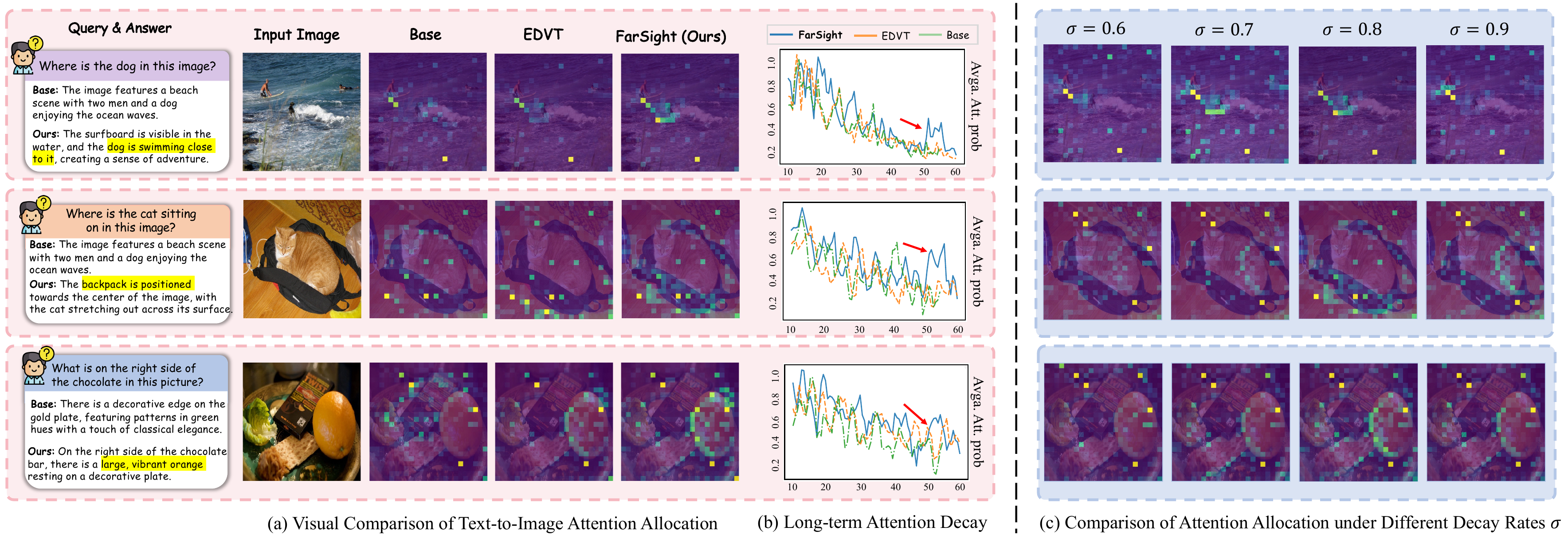} 
  \end{tabular}
  \caption{Qualitative Visualization of FarSight in Image Understanding Task on LLaVA-1.5. (a) Comparison of the average attention allocation to images during text generation among Base (Vanilla MLLMs), EDVT and our FarSight; (b) Visual attention decay across different methods within the generation of 60 text tokens; (c) FarSight's attention distribution on images under varying decay rat $\sigma$. More detailed visualizations of images and videos are provided in Appendix F.}
  \vspace{-0.4cm}
  \label{visual}
\end{figure*}

\subsection{Abalation Study}

\noindent\textbf{Effect of Attention Registers.} 
We experiment with various register-attention values to assess their impact on attention register performance.
As shown in Fig.~\ref{ablation}, our FarSight method improves performance by +6.4\% and +5.4\% on CHAIR$_S$ for LLaVA-1.5 and Video-LLaVA, respectively, significantly outperforming other attention values.
In contrast, the causal masking with -$\infty$ restricts attention allocation in the upper triangular matrix, leading to instability in long-distance dependencies and reduced accuracy. Zero-padding fails to absorb excess attention effectively, increasing the risk of hallucinations during text generation. Although a fixed value of $10^{-3}$ introduces moderate attention absorption, which prevents excessive focus on irrelevant tokens, it still underperforms compared to our method.

\noindent\textbf{Effect of Positional Awareness Encoding.} 
We adopt various positional embedding strategies in the attention layer to assess their impact on the hallucination performance. As shown in Table~\ref{Position}, the baseline RoPE~\cite{su2021roformer}, FixVPE~\cite{Ma_2024_CVPR} and EDVT~\cite{Ma_2024_CVPR} strategies in LLaVA-1.5 and Video-LLaVA result in high hallucination rates. Specifically, RoPE introduces relative positional encoding between visual and text tokens, reducing attention to visual tokens during text generation. Although FixVPE's fixed positional embeddings enhance the consistency of visual information, they are less effective than EDVT's equidistant attention strategy. In contrast, our FarSight significantly improves CHAIR performance by using a progressively diminishing causal mask, retaining attention on earlier tokens (\textit{e.g.,} visual tokens).

\noindent\textbf{Effect of Decay Factor in Attention Registers.} 
We investigate the effect of query sequence lengths on attention decay, as shown in Fig.~\ref{factor}. In both the POPE-R and MSRVTT-QA datasets, MLLMs achieve peak accuracy at a sequence length of 256, with performance starting to decline as the sequence length continues to increase. This can be attributed to the decay factor, which is closely linked to the sequence length. Specifically, as defined in Section~\ref{51} (Implementation Details), the decay factor is influenced by the sequence length and directly affects the rate of attention decay. 
For shorter sequences, the decay factor rises rapidly, limiting the model's ability to capture distant context. Conversely, for longer sequences, the decay factor may initially have a less pronounced effect, but as sequence length increases, attention distribution becomes diluted, increasing decay and information redundancy. A moderate sequence length (\textit{e.g.,} 256) effectively balances the decay factor, maintaining optimal focus on key information and preventing dispersion. 

\noindent\textbf{Quantitative Analysis.}
We visualize the responses and performance of LLaVA-1.5 across different methods and scenarios. Fig.~\ref{visual} (a) shows that FarSight achieves higher accuracy in identifying query-relevant key regions than Baseline and EDVT. This improvement results from its dynamic attention register, which reallocates attention to task-related visual information and reduces attention to irrelevant tokens. The long-term decay curves in Fig.~\ref{visual} (b) show that FarSight maintains strong attention on image tokens in later generation stages, enabled by progressive positional encoding that balances attention between visual and textual tokens throughout the sequence. Fig.~\ref{visual} (c) shows attention distribution under varying decay rates. As the decay rate increases, the model's attention becomes progressively more concentrated, reaching optimal focus at a decay rate of 0.8. However, when the decay rate further increases to 0.9, attention starts to disperse. This indicates the importance of a moderate decay rate for balanced attention.

\begin{figure}[!t]
  \centering
  \begin{tabular}{cc}
    \includegraphics[width=0.95\columnwidth]{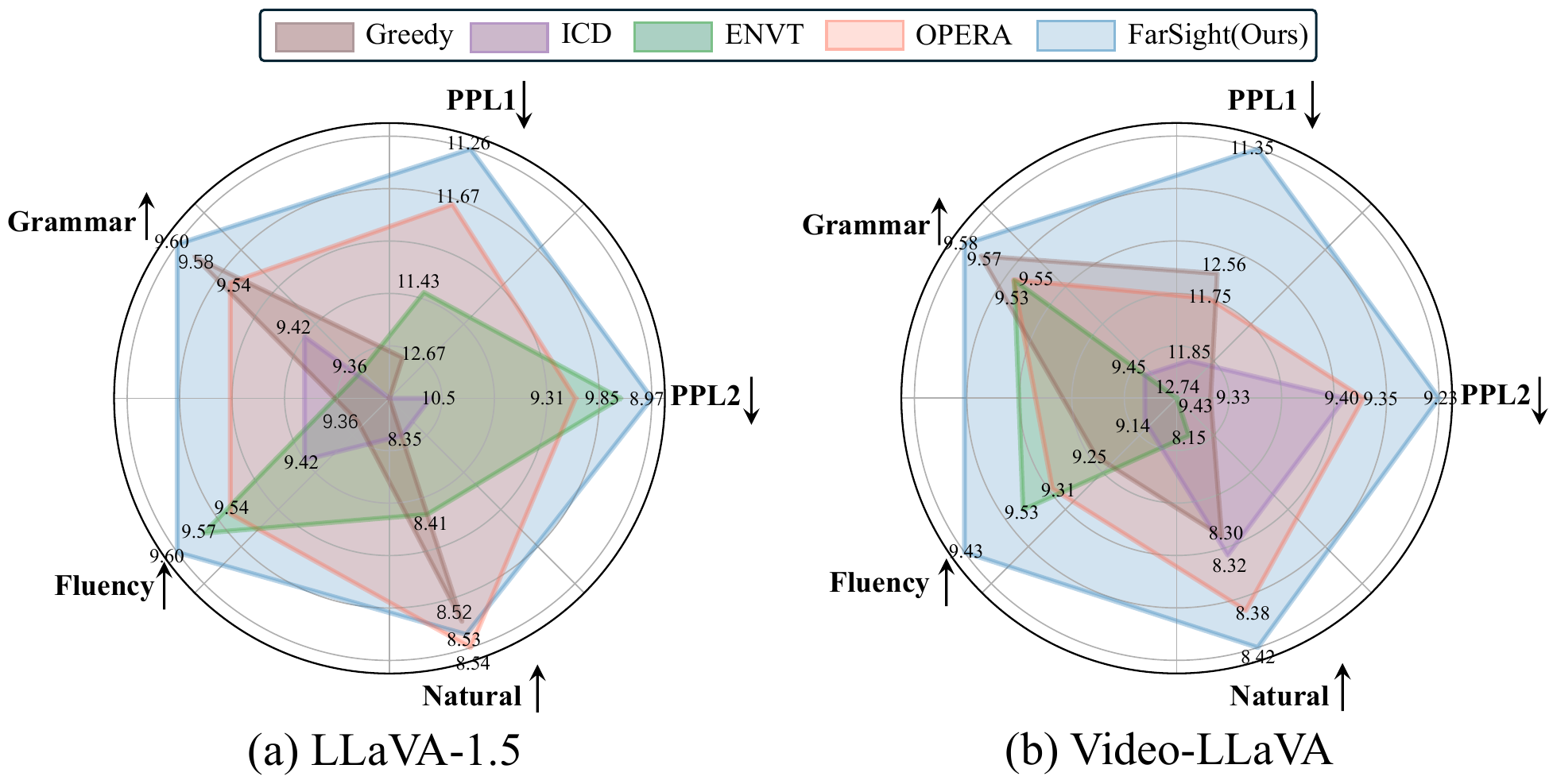} 
  \end{tabular}
  \caption{The average performance is evaluated on a randomly selected set of 600 images from the MSCOCO dataset. PPL$_1$ and PPL$_2$ are calculated using GPT-3.5 Turbo, while the ratings for Grammar, Fluency and Naturalness are provided by GPT-4o.}
  \label{gpt4o}
\end{figure}

\begin{table*}[h]
    \centering
        \caption{Comparison of different MLLMs and FarSight across all image benchmarks. Notably, in the Hallucination Benchmark, lower scores on CHAIR$_{I}$ and CHAIR$_{S}$ indicate better performance, while higher scores are preferable for other metrics.}
    \vspace{-0.15cm}
    \footnotesize
    \resizebox{\textwidth}{!}{
   \setlength{\tabcolsep}{2pt}  
   \setlength{\extrarowheight}{0pt}
   \renewcommand{\arraystretch}{1.1} 
    \begin{tabular}{l|| c c c | c c | c c c c c}
         \midrule
        \rowcolor{mygray}
            \multicolumn{1}{l||}{\cellcolor{mygray}} & \multicolumn{3}{c|}{\cellcolor{mygray}\textbf{Comprehensive Benchmark}} & \multicolumn{2}{c|}{\cellcolor{mygray}\textbf{General VQA}} & \multicolumn{5}{c}{\cellcolor{mygray}\textbf{Hallucination Benchmark}} \\
        \cline{2-11}
        \rowcolor{mygray}
\multicolumn{1}{l||}{\multirow{-2}{*}{\cellcolor{mygray}\textbf{Method}}} & \textbf{MMBench $\uparrow$} & \textbf{LLaVA$^{\mathrm{W}}$} & \textbf{MM-Vet$\uparrow$} & \textbf{VizWiz$\uparrow$} & \textbf{SQA$\uparrow$} & \textbf{CHAIR$_{S}$ $\downarrow$} & \textbf{CHAIR$_{I}$ $\downarrow$} & \textbf{POPE-R$\uparrow$} & \textbf{POPE-P$\uparrow$} & \textbf{POPE-A$\uparrow$} \\
        \midrule
        LLaVA-1.5 & 64.3 & 72.5 & 30.5 & 48.5 & 64.5 & 48.0 & 13.9 & 87.0 & 82.8 & 76.6 \\
        +ICD & 63.1 & 69.7 & 30.4 & 46.9 & 62.8 & 47.7 & 13.6 & 87.9 & 84.0 & 80.2 \\
        +VCD & 63.9 & 70.9 & 29.5 & 43.4 & 63.3 & 46.8 & 13.2 & 87.0 & 83.5 & 78.1 \\
        +OPERA & 64.4 & 72.0 & 31.4 & 50.0 & 64.9 & 45.2 & 12.7 & 88.8 & 82.8 & 79.2\\
        \rowcolor{gray!20} \textbf{+ FarSight (Ours)} & 66.0~\color{blue!60}{({+1.7})} & 74.7~\color{blue!60}{({+2.2})} & 32.5~\color{blue!60}{({+2.0})} & 50.8 ~\color{blue!60}{({+2.3})}& 67.4~\color{blue!60}{({+2.9})} &  41.6~\color{blue!60}{({+6.4})} & 13.2~\color{blue!60}{({+0.7})} & 90.5~\color{blue!60}{({+3.5})} & 86.1~\color{blue!60}{({+3.3})} & 80.4~\color{blue!60}{({+3.8})} \\
        \cdashline{1-11}
        InstructBLIP & 43.4 & 58.2 & 25.6 & 33.4 & 62.1 & 55.6 & 24.2 & 88.7 & 81.3 & 74.4 \\
        \rowcolor{gray!20} \textbf{+ FarSight (Ours)} & 46.5~\color{blue!60}{({+3.1})} & 61.0~\color{blue!60}{({+2.8})} & 27.8~\color{blue!60}{({+2.2})} & 36.0 ~\color{blue!60}{({+2.6})}& 63.4~\color{blue!60}{({+1.3})} & 51.8~\color{blue!60}{({+3.8})} & 23.0~\color{blue!60}{({+1.2})} & 89.5~\color{blue!60}{({+0.8})} & 85.8~\color{blue!60}{({+4.5})} & 76.7~\color{blue!60}{({+2.3})} \\

        Video-LLaVA  & 60.9 & 73.1 & 32.0 & 48.1 & 64.6 & 50.2 & 15.6 & 81.6 & 85.3 & 86.2 \\
        \rowcolor{gray!20} \textbf{+ FarSight (Ours)} & 62.8~\color{blue!60}{({+1.9})} & 74.5~\color{blue!60}{({+1.4})} & 32.8~\color{blue!60}{({+0.8})} & 50.3~\color{blue!60}{({+2.2})} & 66.2~\color{blue!60}{({+1.6})} & 44.8~\color{blue!60}{({+5.4})} & 12.9~\color{blue!60}{({+2.7})} & 83.2~\color{blue!60}{({+1.6})} & 85.8~\color{blue!60}{({+0.5})} & 87.1~\color{blue!60}{({+0.9})} \\

        Chat-UniVi & 56.3 & 70.4 & 28.3 & 46.9 & 59.9 & 52.3 &  16.7 & 85.1 & 69.5 & 64.4 \\
        \rowcolor{gray!20} \textbf{+ FarSight (Ours)} & 59.8~\color{blue!60}{({+3.5})} & 72.6 ~\color{blue!60}{({+2.2})}& 30.7~\color{blue!60}{({+2.4})} & 48.2~\color{blue!60}{({+1.3})} & 62.4~\color{blue!60}{({+2.5})} & 48.9~\color{blue!60}{({+3.4})} & 15.2~\color{blue!60}{({+1.5})} & 87.4~\color{blue!60}{({+2.3})} & 69.7~\color{blue!60}{({+0.2})} & 65.3~\color{blue!60}{({+0.9})} \\
        \midrule
    \end{tabular}}
    \label{image_result}
\end{table*}

\begin{table*}[t]
\centering
\caption{Comparison of different Video MLLMs and FarSight across all video benchmarks. In the Video-Based Text Generation Benchmark, five scores are assessed: \textbf{Cr.} (Correctness of Information), \textbf{Cs.} (Consistency), \textbf{De.} (Detail Orientation), \textbf{Ct.} (Contextual Understanding) and \textbf{Te.} (Temporal Understanding).  
Following Maaz et al.~\cite{Maaz2023VideoChatGPT}, we use the GPT-3.5 Turbo model to assign a relative score to the model outputs, with scores ranging from 0 to 5. See Appendix E for further details.}
\vspace{-0.3cm}
\small
\setlength{\tabcolsep}{10pt}
\renewcommand\arraystretch{1.1}
\resizebox{\linewidth}{!}{
\begin{tabular}{l|| c c| c c| c c c c c}
\midrule
\rowcolor{mygray}
\multicolumn{1}{l||}{\cellcolor{mygray}} & \multicolumn{2}{c|}{\textbf{MSVD-QA}} & \multicolumn{2}{c|}{\textbf{ActivityNet-QA}} & \multicolumn{5}{c}{\textbf{Video-Based Text Generation}} \\
\cline{2-10}
\rowcolor{mygray}
\multicolumn{1}{l||}{\multirow{-2}{*}{\cellcolor{mygray}\textbf{Method}}} & \textbf{Accuracy$\uparrow$} & \textbf{Score$\uparrow$} & \textbf{Accuracy$\uparrow$} & \textbf{Score$\uparrow$} & \textbf{Cr.$\uparrow$} & \textbf{Cs.$\uparrow$} & \textbf{De.$\uparrow$} & \textbf{Ct.$\uparrow$} & \textbf{Te.$\uparrow$} \\
 \midrule
Chat-UniVi  & 64.6 & 3.6 & 43.1 & 3.2 & 2.84 & 2.93 & 2.55 & 3.16 & 2.43 \\
\rowcolor{gray!20} \textbf{+ FarSight (Ours)} & 66.4~\color{blue!60}{({+1.8})}  & 3.5 & 43.7~\color{blue!60}{({+0.6})} & 3.2 & 2.86 & 2.94 & 2.56 & 3.19 & 2.48 \\
Video-LLaVA  & 64.8 & 3.7 & 41.5 & 3.3 & 2.32 & 2.34 & 2.65 & 2.75 & 2.09 \\
\rowcolor{gray!20} \textbf{+ FarSight (Ours)} & 66.2 ~\color{blue!60}{({+1.4})}& 3.6 & 42.0 ~\color{blue!60}{({+0.5})}& 3.5 & 2.43 & 2.38 & 2.93 & 2.84 & 2.14 \\
VILA & 72.6 & 4.0 & 50.2& 3.3 & 3.14 & 3.40 & 2.71 & 3.43 & 2.58 \\
\rowcolor{gray!20} \textbf{+ FarSight (Ours)} & 74.5~\color{blue!60}{({+1.9})} & 4.2 & 51.4    ~\color{blue!60}{({+1.2})}& 3.6 & 3.18 & 3.52 & 2.73 & 3.45 & 2.60 \\
Video-LLaMA2   & 70.9 &  3.8  & 49.9 & 3.3 & 3.13 & 3.23 & 2.70 & 3.42 & 2.45 \\
\rowcolor{gray!20} \textbf{+ FarSight (Ours)} & 73.8~\color{blue!60}{({+2.9})} & 3.9 & 50.4~\color{blue!60}{({+0.5})} & 3.6 & 3.26 &  3.32 & 3.21 & 3.50 & 2.47 \\
\midrule
\end{tabular}}
\label{video_result}
\end{table*}

\subsection{Comparison to State-of-the-Arts}
\noindent\textbf{GPT-4o Assisted Evaluation.}
To comprehensively evaluate the overall quality of generated text, we employ the PPL (Perplexity) metric and utilize GPT-4o to assess the grammar, fluency, and naturalness of the text. We randomly select 600 images from the MSCOCO dataset and perform validation using the LLaVA-1.5 and Video-LLaVA. As demonstrated in Fig.~\ref{gpt4o}, FarSight consistently preserves the quality of the generated text across multiple dimensions.

\noindent\textbf{Image Benchmarks Evaluation.} 
To evaluate the image understanding, we compare models with the FarSight extension against several decoding methods, including ICD~\cite{wang2024mitigating}, VCD~\cite{leng2024mitigating} and OPERA~\cite{huang2024opera}, as shown in Table~\ref{image_result}. Integrating FarSight as a plugin into LLaVA-1.5 results in an average improvement of +2\% in the Comprehensive and General VQA tasks. It also achieves significant gains in hallucination metrics, with CHAIR$_{S}$ and POPE-P scores increasing by +6.4\% and +3.3\%, respectively. These results indicate that FarSight is effective at reducing hallucinations in both structured and unstructured environments. Furthermore, the benefits of FarSight extend beyond the LLaVA-1.5 model, as other models also experience considerable enhancements, especially in hallucination evaluation tasks, with the CHAIR$_{S}$ metric increasing.

\noindent\textbf{Video Benchmarks Evaluation.} In Zero-Shot Video Question Answering Tasks, FarSight achieves significant improvements over video MLLMs across three key benchmark datasets. As shown in Table~\ref{video_result}, on the MSRVTT-QA dataset, our method delivers an average accuracy gain of +3\% across multiple models, reaching a peak accuracy of 68.9\%. On MSVD-QA and ActivityNet-QA datasets, FarSight improves accuracy by +2\% and +0.7\%, respectively, demonstrating consistent enhancements across different video contexts and question types. Moreover, in Video-Based Text Generation, the integrated model outperforms the baseline MLLMs across five critical dimensions. 

\section{Conclusion}
In this work, we analyze the self-attention token propagation patterns, revealing two main causes of hallucinations in MLLMs: attention collapse and positional information decay. To mitigate them, we present FarSight, a plug-and-play decoding strategy that reduces interference from outlier tokens and enhances in-context inference. The core of our method is effective token propagation, which is achieved by optimizing the causal mask with attention registers and a diminishing masking rate. Extensive experiments on both image and video tasks have shown that the proposed method outperforms existing state-of-the-art methods, and the ablation study has revealed the effectiveness of our FarSight.

\section{Acknowledgment} We also thank the remaining co-authors of the CVPR 2025 version of this work for their contributions to the original publication. We would like to thank Zile Huang, Haochen Xue, and Ziyang Chen for their support in dataset collection and preprocessing. We are also grateful to Sijin Zhou, Wenxue Li, Yulong Li, Wenxuan Song, and Wei Feng for their contributions to model implementation, training, and experimental validation. We would also like to sincerely thank Qingyu Yin for his technical guidance and insightful suggestions related to causal masking design.

\appendix

\section{Implementation details for Figure 2}

We randomly select 500 samples from the CHAIR dataset and conduct a Snowball Hallucination analysis. The process is outlined as follows:
\begin{enumerate}
    \item Input the original prompts (\textit{i.e.,} ground truth descriptions of the images) and the text generated by MLLMs into GPT-4o. GPT-4o is prompted to perform a sentence-by-sentence analysis of the generated text, examining whether each statement is consistent with the original prompt.
    \item The position and specific content of the first occurrence are recorded. Subsequent hallucinations in the text are analyzed to determine whether they derive from the initial hallucination. If subsequent hallucinations are logically dependent on the initial hallucination (\textit{e.g.,} extrapolated or inferred based on incorrect information), they are classified as snowball hallucinations; otherwise, they are categorized as independent hallucinations.
    \item Quantify the ratio of snowball hallucinations to independent hallucinations to evaluate the factual accuracy of the generated text. The detailed prompt is provided at the end of the Appendix.
\end{enumerate}

\subsection{Image Heatmap Visualization}
We analyze the responses and performance of Image and Video MLLMs across various tasks and scenarios. Fig.~\ref{video_heatmap} and~\ref{img_heatmap} illustrate the MLLMs' attention to visual information during the answer generation process. Visualization results demonstrate that, compared to baseline methods, the model achieves higher attention accuracy for image-related queries, highlighting its capability to dynamically focus on task-relevant visual features. This improvement is attributed to the dynamic attention register mechanism, which prioritizes key visual regions while effectively reducing interference from irrelevant tokens. Notably, this phenomenon is not limited to image MLLMs but also exhibits strong adaptability in video MLLMs, further validating the broad applicability of this mechanism.

\subsection{Video Heatmap Visualization}
\begin{figure}[!h]
  \centering
  \begin{minipage}{\textwidth}
    \centering
\includegraphics[width=0.9\linewidth]{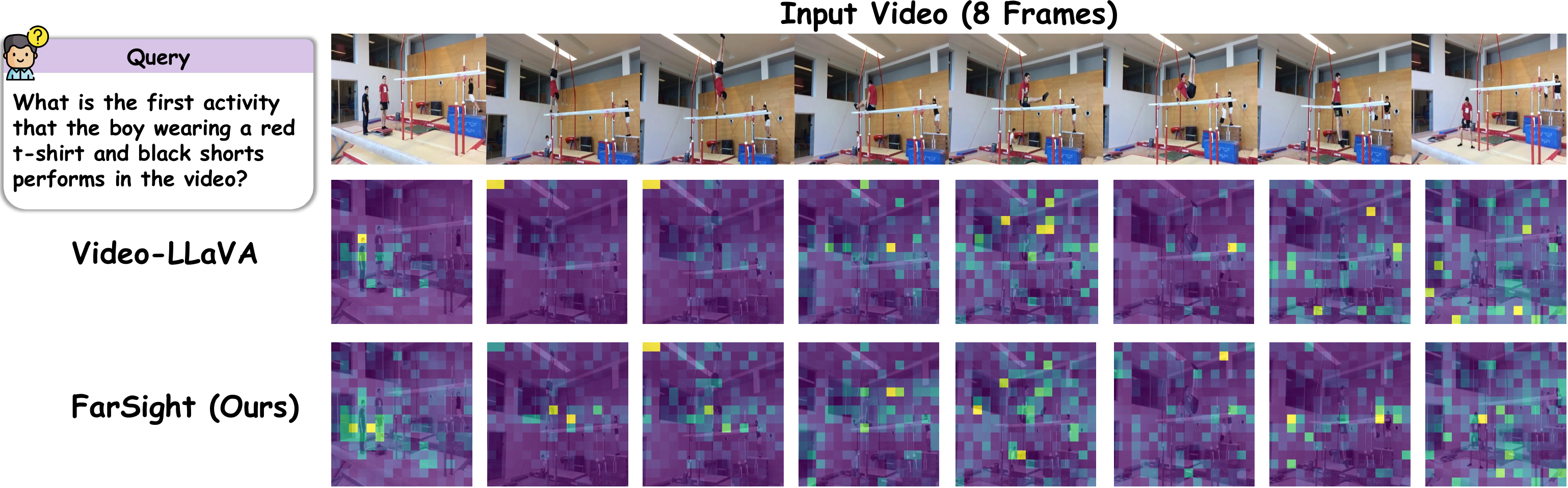}
    \caption{Qualitative Visualization Example 1. 
The attention distribution of Video-LLaVA is relatively scattered and inconsistent, failing to focus on the boy wearing a red t-shirt and black shorts mentioned in the query. Instead, the attention is dispersed across multiple areas, indicating difficulties in isolating task-relevant regions. In contrast, FarSight exhibits more focused and consistent attention, clearly targeting the boy and his key features.}
    \label{video1}
    \vspace{1cm}   \includegraphics[width=0.9\linewidth]{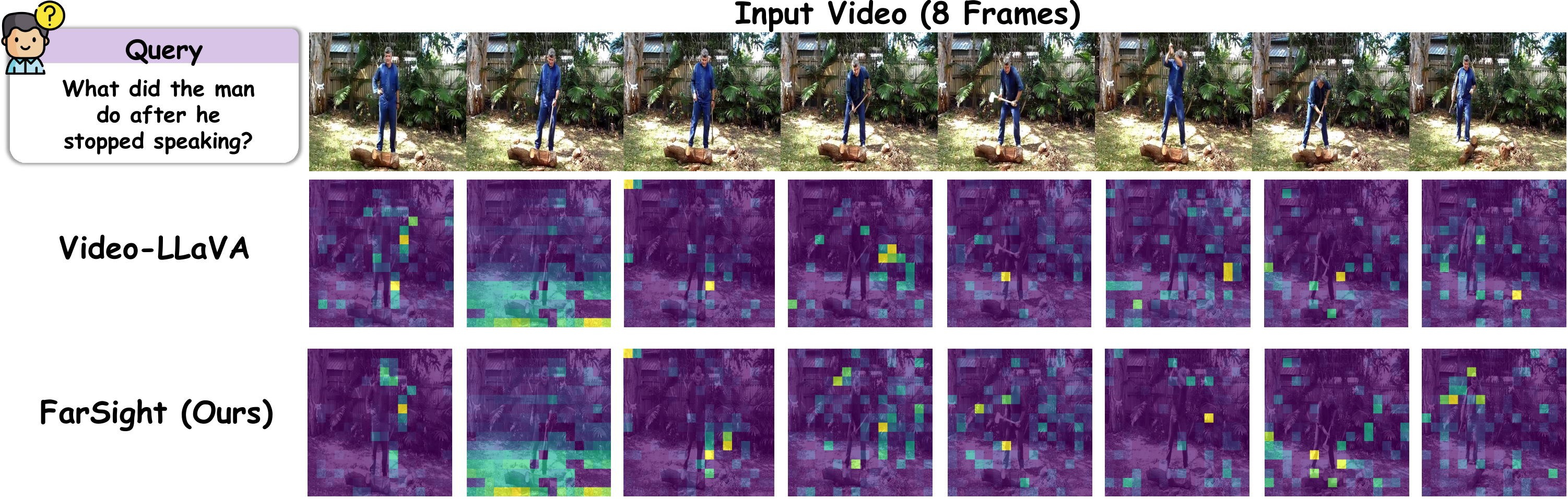}
    \caption{Qualitative Visualization Example 2. The attention distribution of Video-LLaVA fails to consistently focus on the man and his actions, with attention often directed toward other areas of the scene, such as the background or non-essential objects. In contrast, the attention distribution of FarSight is significantly more concentrated, accurately targeting the man’s key body parts, such as his hands and the areas where he interacts with objects.}
    \label{video2}

    \vspace{1cm}

\includegraphics[width=0.9\linewidth]{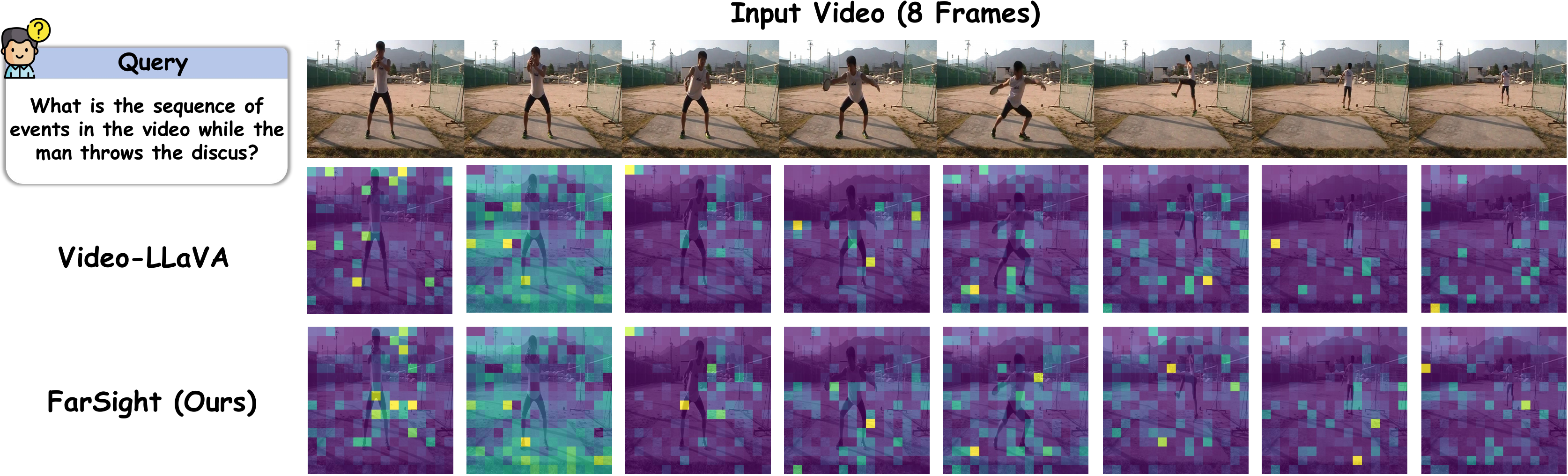}
    \caption{Qualitative Visualization Example 3. 
Video-LLaVA's attention lacks sufficient temporal coherence. In contrast, FarSight demonstrates precise and consistent focus on the man and his discus-throwing actions, successfully capturing the complete sequence of events.}
    \label{video3}
  \end{minipage}
\end{figure}

\begin{figure*}[htbp] 
  \centering
  \begin{tabular}{cc}
    \includegraphics[width=1.6\columnwidth]{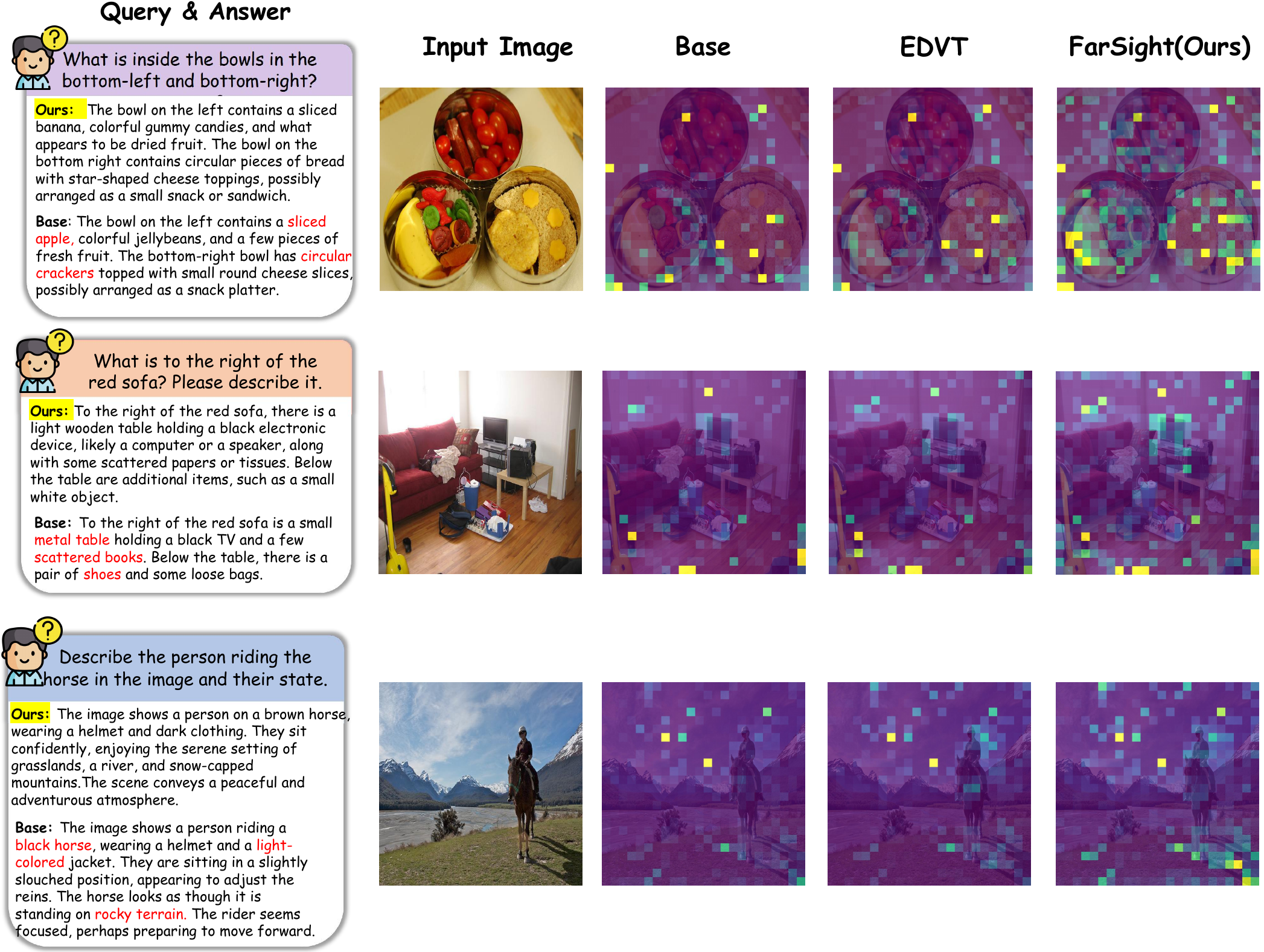}
  \end{tabular}
    \caption{LLaVA-1.5 Qualitative Visualization.}
  \label{video_heatmap}
\end{figure*}

\begin{figure*}[htbp] 
  \centering
  \begin{tabular}{cc}
    \includegraphics[width=1.6\columnwidth]{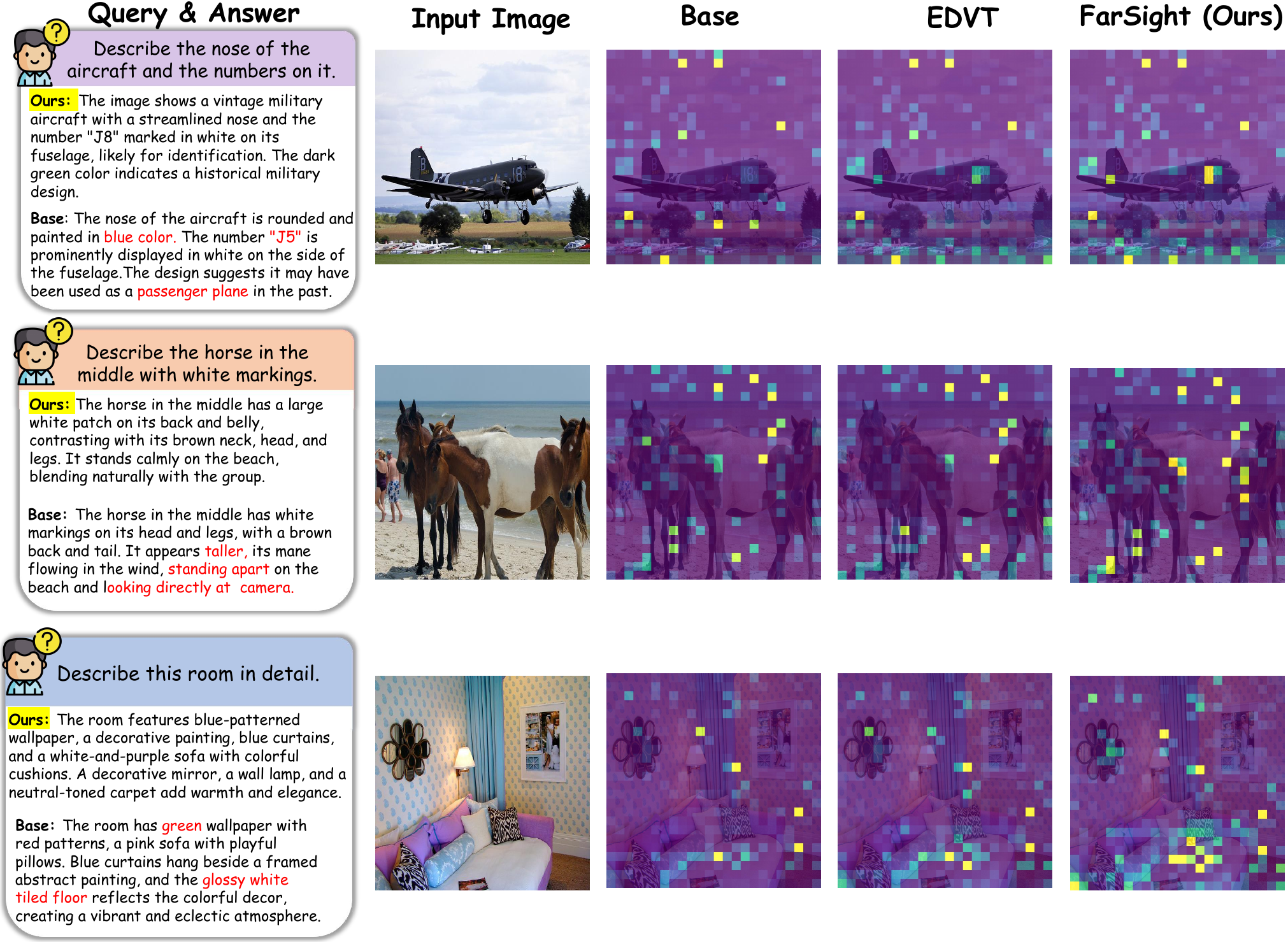}
  \end{tabular}
    \caption{Video LLaVA Qualitative Visualization.}
  \label{img_heatmap}
\end{figure*}

Fig.~\ref{video1}-\ref{video3} illustrate the attention distribution of Video-LLaVA across three video scenarios, focusing on how the generated text aligns with visual information. Since Video-LLaVA natively supports either 8 or 16 frames, we adopt the 8-frame extraction method in our experiments to ensure efficient inference. The visualizations demonstrate that FarSight excels in capturing complex spatiotemporal information, such as human actions and scene details. While Video-LLaVA also maintains a relatively strong attention to visual elements, its focus becomes increasingly dispersed and less concentrated over time. For example, as shown in Fig.~\ref{video2}, which depicts a scene of a man chopping wood, both Video-LLaVA and FarSight perform comparably in the earlier frames, adequately capturing the man’s position and actions. However, as the temporal span increases toward the final three frames, Video-LLaVA exhibits reduced attention to the core features, shifting its focus to surrounding environmental elements. In contrast, FarSight consistently concentrates on the man’s chopping actions, effectively identifying the key visual details.

This phenomenon can be attributed to the application of FarSight's progressive positional encoding, which effectively maintains attention allocation to early visual tokens during sequence generation. Unlike traditional positional encoding strategies, which often lead to a gradual decline in attention to earlier frames as the sequence progresses, progressive positional encoding dynamically adjusts positional weights to ensure balanced attention distribution across the temporal span. This strategy enables the model to concurrently focus on both earlier and later visual tokens, thereby maintaining consistent attention to critical spatiotemporal features throughout the video. This design resolves the issue of diminishing attention to earlier tokens while enhancing the MLLM's ability to integrate and prioritize task-relevant information across frames, improving its performance and reliability in complex spatiotemporal scenarios.

\clearpage

\clearpage

{
    \small
    \bibliographystyle{ieeenat_fullname}
    \bibliography{main}
}


\end{document}